# A study on the deviations in performance of FNNs and CNNs in the realm of grayscale adversarial images

Durga Shree N[1,2]
durgashree.n2020@vitstudent.ac.in

Steve Mathew D A[1,2]
stevemathew.da2020@vitstudent.ac.in

Chiranji Lal Chowdhary[1]*
chiranji.lal@vit.ac.in

[1]Department of Software and Systems Engineering,

School of Information Technology and Engineering,

Vellore Institute of Technology, Vellore, India - 632014

[2]Data Science and Applications,

Indian Institute of Technology Madras, Chennai, India - 600036

Corresponding Author*

**Abstract**

Neural Networks are prone to having lesser accuracy in the classification of images with noise perturbation. Convolutional Neural Networks, CNNs are known for their unparalleled accuracy in the classification of benign images. But our study shows that they are extremely vulnerable to noise addition while Feed-forward Neural Networks, FNNs show very less correspondence with noise perturbation, maintaining their accuracy almost undisturbed. FNNs are observed to be better at classifying noise-intensive, single-channeled images that are just sheer noise to human vision. In our study, we have used the hand-written digits dataset, MNIST with the following architectures: FNNs with 1 and 2 hidden layers and CNNs with 3, 4, 6 and 8 convolutions and analyzed their accuracies. FNNs stand out to show that irrespective of the intensity of noise, they have a classification accuracy of more than 85%. In our analysis of CNNs with this data, the deceleration of classification accuracy of CNN with 8 convolutions was half of that of the rest of the CNNs. Correlation analysis and mathematical modelling of the accuracy trends act as roadmaps to these conclusions.

1. Introduction

Adversarial examples are images that are created by adding noise that is specifically curated to fool machine learning models [1]. These have posed a serious threat to image classification algorithms and systems in recent times. The primary concern with regard to these adversarial examples is that they are not distinguishable (the noise added is not profound) from the benign images to the human eye. The technique in adversarial machine learning where the model is fooled by testing against adversarial images is called evasion attack. Previous studies train the model using adversarial input with an attack step size of upto 16 to introduce the model to adversarial data [2]. This training teaches the model to expect changes at the pixel level and act accordingly while encountering adversarial images in the test set. In our study, we train using the benign dataset and test the learning capabilities of the model using images severely exposed to noise distortion. This is to evaluate the model's robustness if it will be exposed to adversarial images in real-time. This nature of training is identical to what is most probably possible to happen while attempting to tamper an ML model.

In this study, we have used Feed Forward Neural Networks and Convolutional Neural Networks and compared their accuracies in the realm of adversarial test sets. Feed Forward Neural Networks and Convolutional Neural Networks view the image from different perspectives. While the strategy of CNNs is highly appreciable on benign datasets, the stability and robustness of FNNs is observed to be commendable on noisy, single-channeled images. FNN deals with individual pixels to derive patterns for each output bin while CNNs emphasize on the concept of utilizing the data of pixels in the proximity of a pixel of interest (local spatial coherence) to understand the associations between them[3][4]. The arrangement of the animal visual cortex, whose individual neurons are organized in such a way that they respond to overlapping portions of the visual field, served as the inspiration for CNN's connection network. Though both of them are very capable when it comes to image classification tasks, CNNs can only understand the elements in an image by identifying the proximal positions of the relevant pixels. The relative position of an object in the image does not affect CNNs but FNNs are very sensitive to the position of the object of interest in the image.

Previous works have been focussed on comparing and contrasting Deep Belief Networks (DBN) and CNNs in adversarial settings. It was found that DBNs performed better than CNNs for adversarial examples. The downside of CNNs was due to the strong inductive bias assumption which is an attribute of the working of CNN [5]. Alterations have been made to the CNN architecture to come up with models that improve the robustness of CNNs against adversarial examples [6]. In a unique study, CNN architecture has been altered in such a way that it is able to denoise an image [7]. Previous studies have been engaged in finding the appropriate activation function for the hidden layer of an FFN. Laudani et al. also proposed a method to change a network configuration between various activation functions without affecting the network mapping capabilities [15].

There are many real-life scenarios where adversarial images can result in a huge material loss and can lead to compromised decisions. Adversarial Images can fool medical diagnosis systems where something which is malignant could be incorrectly classified benign or vice-versa [18]. Usually, the pixels are altered to fool machine learning models. A new approach of adding an adversarial patch that can influence the decision of a model is also a viable adversarial attack [19]. Adversarial attacks are not just reserved to image dataset classifiers. Time series classifiers, speech recognition systems, video processing systems, object detectors, etc can also be vulnerable to adversarial attacks [20]. Quality, safety and security monitoring AI models are at risk of being manipulated.

Neural Networks have a standard approach of learning from image datasets. Initially, neural networks assign random weights to the weight matrices that attempt to establish relationships between any two layers. Training is essentially just a process to adjust these weights into meaningful values that capture features of the training dataset. Gradient Descent method of the training process is used to find a local minimum of the loss function. Loss function represents the loss incurred due to insufficient adjusting of the weights [11]. The weights at a particular instance of time can be adjusted by taking the gradient of the loss function at that point and by stepping in the negative direction of the gradient. Cross-entropy [12] is a differentiable function which accounts for providing feedback towards stepwise improvement of the model by assigning higher probability to the correct label in order to reduce loss.

The captured features do not sufficiently support the classification of noisy images due to deviations from expected patterns of test inputs at pixel level. In real-life scenarios it is possible that the model will come across images that are distorted, noise-intensive and misleading. Such images can also be created computationally. The generation of adversarial images follows a procedure opposite to that of gradient descent. Differentiating the loss function with respect to parameters to decrease the loss on the sample is done to reach the minimum (at least local) in gradient descent. Similarly, differentiating the loss function with respect to the input data to modify the input data such that the expected loss of the model increases in the sample data generates the required sample of adversarial images. This significant relationship between model-training and model-fooling is one of the interesting aspects of exploring adversarial robustness. Fast Gradient Sign Method, FGSM, a method proposed by Goodfellow, et al, to generate adversarial inputs [9] is precisely the same as doing one gradient ascent step, with the exception that we fix the perturbation on each pixel to be a constant size - epsilon, which ensures that no pixel in the adversarial example differs from the original picture by more than the value, epsilon.

In order to evaluate the robustness of neural networks against noise perturbation in image recognition, we add noise to normal images. Some of the techniques that are used to add noise to the images are: Fast Gradient Sign Method, One-step target class methods, Basic iterative method, Iterative least-likely class method etc., [2]. There are also numerous methodologies applied to defend a model from adversarial attacks. The two common approaches are, increasing robustness of machine learning models and detecting adversarial attacks. To increase the robustness of machine learning models, the models can be trained with adversarial examples. The other methods include defensive distillation, random resizing and padding, stochastic activation pruning, total variance minimization and quilting, thermometer encoding, adversarial logit pairing, etc. To detect adversarial attacks, the initial approaches used were principal component analysis, softmax, and reconstruction of adversarial images. Other recent techniques include feature squeezing, adversary detector networks, reverse cross-entropy, kernel density and bayesian uncertainty estimates [14].

In the following sections, we perform experiments to compare the classification accuracy of different neural networks and derive interpretations from the results obtained to further understand the behavior of the models under study for both benign and adversarial test samples.

## 2. Methods

### 2.1 Architecture of FNNs used in the study

In this architecture, no neuron in the output layer acts as an input to a preceding layer or the same layer. Two single-layer feedforward networks with hidden layers of sizes 32 and 256 neurons are constructed. Another feedforward neural network which consists of two hidden layers of 256 and 32 neurons is constructed for the study.

The shape of each sample image in the MNIST dataset is 28x28. There are 784 input neurons which represent the 784 pixels. For instance, if we consider the model with 32 neutrons in its hidden layer, the size of the input layer would be 1x784 and the dimensions of the weight matrix for the hidden layer would be of the size 784x32. After the data passes through the hidden layer, the weight matrix for the output layer would be of the size 32x10. When we multiply the matrix of the input layer with that of the weight matrix between the input and hidden layer, we will arrive at a matrix of size 1x32 and when the same is multiplied with the weight matrix between the hidden and the output layer, we would arrive at a matrix of size 1x10 which denotes the output of the neural network. The final 10 output values tell the probability of the test image belonging to any one of the output bins. ReLU is applied after each of the layers to keep the relationships between the input and output non-linear. Softmax function is applied after the last layer to assign probabilities of classification to all the bins.

### 2.2 Architecture of CNNs used in the study

In the CNNs developed, a kernel of size 3 is used to traverse over the pixels in the image. A kernel is essentially a filter which is used for feature extraction. Several kernels together form a convolution that acts as the feature repository at a particular stage in the feature extraction process. A padding of 1 is used to make sure that the dimensions of the image remain the same even after sliding the kernels. After each of the convolutions, the activation function ReLU is used before the next convolution is performed. After every two convolutions, max-pooling is done to condense the information into smaller matrices [13].

A kernel performs the task of extracting the low-level features. Once max-pooling is applied on the convolution of low-level features, other high-level features are extracted through subsequent filters (hierarchical feature learning). Stride is the number of pixels by which the kernel is shifted each time during the traversal. When the value of the stride is very high, then the kernel 'hops' leaving many pixels in between the hops. For instance, when the kernel size is set to 3 and the stride is set to 3, then each pixel will be traversed exactly once. If the value of stride is set low, then the resolution of the filtered image will be high due to application of multiple slides of different parts of the kernel on the same section of matrix in the convolution phase.

**2.3 Experimental Background**

In this study, we have considered the MNIST dataset which is a curated collection of hand-written image datasets in grayscale (single-channeled) with 50000 training images and 10000 test images. FNNs were trained and modeled for 100 epochs and CNNs were trained for a maximum of 50 epochs. The training was continued until the accuracy of the model flatlined.

The method used to generate adversarial examples is the Fast Gradient Sign Method [9] which "linearizes the cost function to obtain an optimal max-norm constrained perturbation". FGSM focusses on adding noise whose direction is the same as the cost function's gradient in accordance to the data.

The activation function which we have used in the study is the Rectified Linear Unit (ReLU) [10]. The output of the hidden layer and the inputs have a linear relationship ie., each element of the output of the hidden layer is the product of the weights and the elements from the input layer. Hence, a linear function of weights has been established. This makes the hidden layers capable of only capturing the linear relationships between the input layer and the output layer. On application of ReLU, being a nonlinear function, helps in the process of activation by increasing the possibilities of capturing relationships that are nonlinear.

Computation of loss is done using the cross-entropy method. Minimization of the computed loss by administering suitable weight adjustments is implemented using the Gradient Descent Method.

A learning rate of 0.01 was used on FNNs throughout the training process. The size of the hidden layer for the FNN model with one hidden layer was set to 256. The size of the hidden layers for the FNN model with two hidden layers was set to 256 and 32. Learning rates ranging between 0.001 and 0.0000001 were used based on the speed of reaching the local minima (minimum loss). To speed up the local minimum reachability, learning rates were decreased gradually.

## 3. Results and Discussions

### 3.1 Performance of Models on benign dataset:

CNNs are widely used for image classification and known for their accuracy in effectively classifying images with multiple channels. But FNNs are not as popular as the former in the realm of image classification. This is because CNNs take into account the relative position of pixels when performing feature extraction. But in FNNs, the relative positions are not taken into account and are just seen as a set of pixels.

| Models | FNN(1 - hidden layer) | FNN(2 - hidden layers) | CNN(3 - convolutions) | CNN(4 - convolutions) | CNN(6 - convolutions) | CNN(8 - convolutions) |
|---|---|---|---|---|---|---|
| Accuracies | 96.20% | 97.70% | 99.47% | 99.52% | 99.47% | 99.45% |

*Table 1. Classification accuracies of the models for benign test images*

Here, FNNs show comparable accuracy with CNNs in classification due to the simplicity of the dataset. In the dataset of interest (MNIST), each sample image is gray-scaled and is composed of 784 pixels (28 X 28). The minimal resolution of sample images is not comparable with those found in real-life scenarios. Owing to this lesser number of pixels and single-channeled input, FNNs were capable of classifying these images to a good degree of accuracy and so FNNs were considered in this study to be compared with the CNNs.

All models show acceptable accuracy in classification of benign images. Increasing the number of layers or convolutions had no effect on accuracy in the respective models. The models showed a very fast rate of learning owing to the smaller number of pixels in each training image and single-channel input. The models showed notably high accuracies from the first epoch itself.

**3.2 Influence of the size of hidden layer on accuracy in FNN with one hidden layer**

When the size of the hidden layer is increased from 32 to 256, we could observe a massive improvement in the accuracy at which the model classifies adversarial images. The model which has 32 neurons in the hidden layer shows a decline in the accuracy as the attack step size is increasing and its accuracy for minimal noise perturbation (epsilon = 0.1) is also not satisfactory. But the one with 256 neurons in the hidden layer has a standard deviation close to 1 in its accuracy with its arithmetic mean being 88%. But this increase from 32 neurons to 256 did not have a big impact on the benign test examples but rather increased the accuracy of classifying adversarial images with a large margin.

| Attack Step Size (Epsilon) | Size of hidden layer = 32 | Size of hidden layer = 256 |
|---|---|---|
| 0.1 | 39.11% | 88.69% |
| 0.2 | 24.54% | 88.23% |
| 0.3 | 25.21% | 87.03% |
| 0.5 | 23.44% | 88.97% |
| 0.75 | 19.73% | 88.51% |
| 1 | 18.80% | 88.61% |
| 1.5 | 17.57% | 88.87% |
| 2 | 19.99% | 88.08% |
| 5 | 16.80% | 88.97% |
| 10 | 24.15% | 86.45% |
| 16 | 21.23% | 86.20% |

*Table 2. Classification accuracies of FNN with 1 hidden layer of sizes 32 and 256 against different intensities of noise*

As the size of the hidden layer increases, feature extraction is effective. This can be reasoned out by the presence of a bigger weight matrix between the preceding layer and the current hidden layer of interest. These effectively extracted features make the model robust to noisy input.

**3.3 Performance of Models on Adversarial dataset:**

A significant amount of deviation is observed in the performance of CNNs from the previous test result on benign dataset. Feedforward Neural Networks show a high and steady accuracy to all the tested range of attack step sizes. The performance of FNNs with one hidden layer of 256 neurons is commendable. Adding another layer of size 32 has no significant effect on improvement in accuracy. A trend of decrease in accuracy is observed in every other model, but FNNs show no trend of variation. The accuracy ranges around 88% and 90% for the two models with just 2% of deviation within the tested attack step sizes.

|  | Classification Accuracies | | | | | |
|---|---|---|---|---|---|---|
| Attack Step Size (Epsilon) | FNN(1 hidden layer) | FNN(2 hidden layers) | CNN(3 convolutions) | CNN(4 convolutions) | CNN(6 convolutions) | CNN(8 convolutions) |
| 0.1 | 88.69% | 94.02% | 94.50% | 90.15% | 94.10% | 96.47% |
| 0.2 | 88.23% | 91.05% | 70% | 65.79% | 75.50% | 83.76% |
| 0.3 | 87.03% | 87.66% | 37.50% | 38.66% | 51.00% | 79.11% |
| 0.5 | 88.97% | 91.35% | 14.50% | 20.19% | 23.02% | 65.15% |
| 0.75 | 88.51% | 88.58% | 19.20% | 14.65% | 14.93% | 38.14% |
| 1 | 88.61% | 89.47% | 16.39% | 15.67% | 14.29% | 40.38% |
| 1.5 | 88.87% | 89.37% | 15.24% | 20.44% | 11.15% | 38.08% |
| 2 | 88.08% | 90.66% | 19.56% | 16.79% | 14.56% | 30.04% |
| 5 | 88.97% | 88.79% | 12.61% | 11.58% | 19.29% | 32.10% |
| 10 | 86.45% | 89.30% | 17.40% | 13.83% | 17.80% | 31.21% |
| 16 | 86.20% | 91.74% | 19.17% | 10.77% | 18.48% | 16.73% |

*Table 3. Classification accuracies of networks constructed with different intensities of noise perturbation*

The probability of guessing a number right without any training is 10% (due to the presence of 10 possible classes each denoting a number). CNNs with 3, 4 and 6 convolutions show no better performance than guessing as epsilon increases beyond a threshold. This observation can be attributed to the perplexed state the model is in due to immense noise accumulation. Kernels act as filters to extract various features of an image at various levels [8]. Huge amounts of distortion from the test image do not give information about features the kernel is expecting to extract. Since the model is bound to classify the image into one of the ten bins, it classifies it based on what it considers is the nearest fit. But the information using which the model arrives at this conclusion is not strongly cemented on clear features extracted as it is done for benign images.

**3.4 Correlation between the Accuracies shown by various models for tested epsilons ($r^2$):**

| Correlation of Accuracies | FNN(1 hidden layer) | FNN(2 hidden layers) | CNN(3 convolutions) | CNN(4 convolutions) | CNN(6 convolutions) | CNN(8 convolutions) |
|---|---|---|---|---|---|---|
| FNN(1 hidden layer) | 1 | 0.09407 | 0.09382 | 0.18296 | 0.07111 | 0.27618 |
| FNN(2 hidden layers) | | 1 | 0.61470 | 0.58903 | 0.53898 | 0.35413 |
| CNN(3 convolutions) | | | 1 | 0.98765 | 0.97756 | 0.82924 |
| CNN(4 convolutions) | | | | 1 | 0.97819 | 0.88976 |
| CNN(6 convolutions) | | | | | 1 | 0.89334 |
| CNN(8 convolutions) | | | | | | 1 |

*Table 4. Correlations between the trend of decline of classification accuracies of networks for adversarial images*

The correlation between the range of accuracies of FNN (1 - hidden layer) and FNN (2 - hidden layers) cannot be commented on because the accuracies of the models do not follow a trend of inclination or declination. The accuracies remain approximately steady for the entire range of noise addition.

The correlation among all possible combinations of the analyzed CNNs are high and it shows that the trend of decline of their accuracies across the range of attack step sizes are similar. Among them, the correlation between the CNNs with 3, 4 and 6 convolutions are very high and their trends of decline are identical to each other.

From the above table, we can infer that the accuracy is the maximum for CNN with 8 convolutions when the attack step size is very low (in our case, 0.1) even when compared to other FNN models. Other CNN models don't correlate with CNN with 8 hidden layers as much as they do among themselves. This slight decline in correlation is due to slightly better performance of CNN (8 convolutions) when compared with other CNNs when the attack step size is increasing beyond 0.5.

**3.5 Visual representation and interpretation with respect to accuracies**

The benign images when added with the carefully curated noise resulted in the below adversarial images. The attack step size (epsilon) denotes the intensity of noise added to the image.

| Epsilon: 0 (Benign image) | Epsilon: 0.1 | Epsilon: 0.2 | Epsilon: 0.3 |
|---|---|---|---|
| 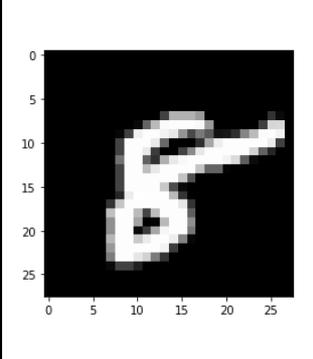 | 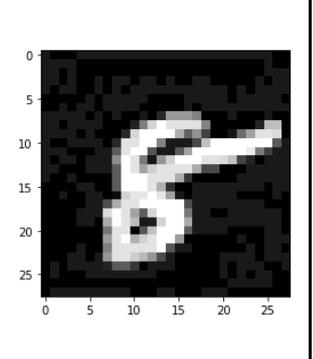 | 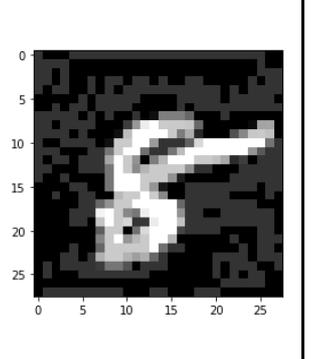 | 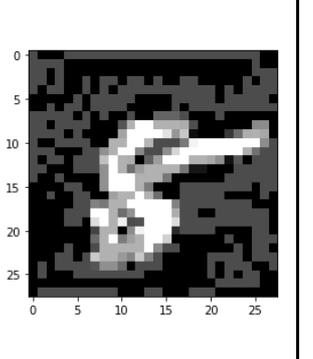 |

***Table 5. Adversarial examples with low enough intensities to be comprehended by human vision***

In the above images, we can see that the number that is in the image is evident to the naked eye. But the addition of noise is also not unnoticeable. The minimal noise added is also noticeable in this dataset because of the lower resolution of the images.

| Epsilon: 0.5 | Epsilon: 0.75 | Epsilon: 1 |
|---|---|---|
| 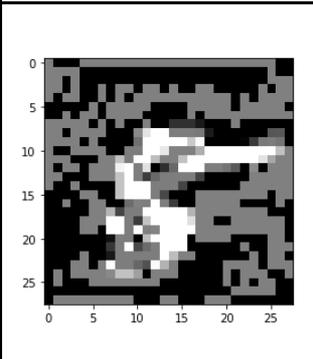 | 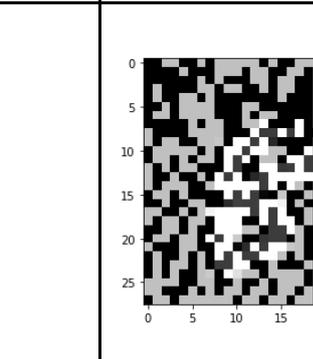 | 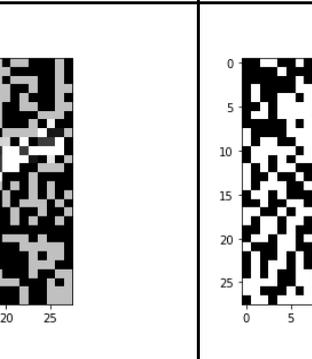 |

***Table 6. Distorted Adversarial Images with high enough noise intensities to conceal the number from visual perception***

In the above images, the concentration of noise is so high such that the number in the image is not identifiable to the human eye. When the concentration of the noise reaches an attack step size of 1, the image just becomes sheer noise and no observation can be drawn from the image for a human.

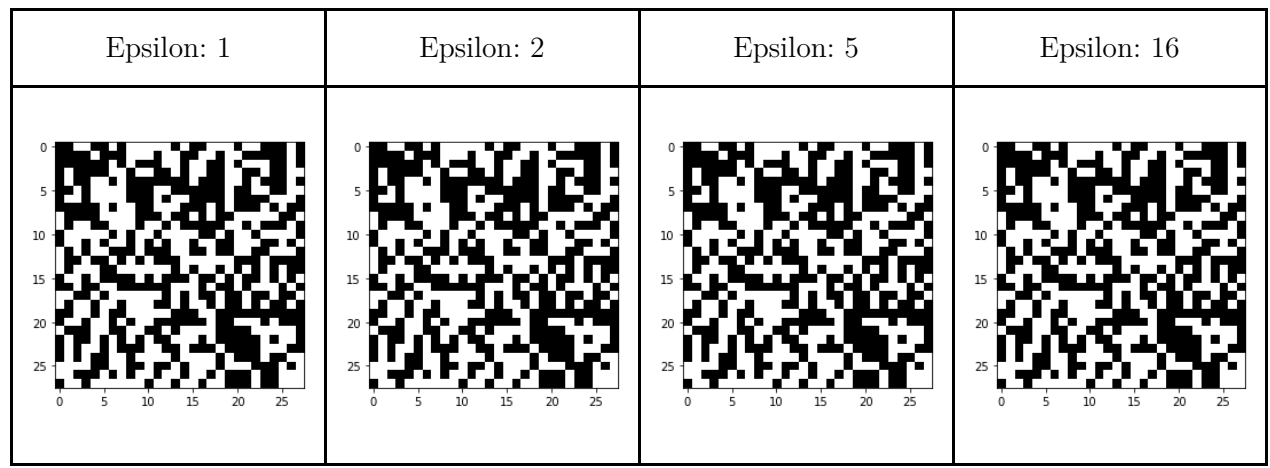

*Table 7. Adversarial Images with noise perturbations that result in the same image for higher degrees of intensities.*

These levels of attack step-size do not have such huge influence in distortion in real-life multi-channeled images. Since the image resolution of this dataset is quite minimal (28x28), these attack step sizes have a drastic effect on the image.

This attack step size of 1 serves as a breaking point for this dataset such that any amount of noise above this limit is resulting in an image that looks the same. This could act as a potential point that challenges the robustness of the CNN models towards adversarial noise. Just as it is evident to the human eye that no peculiar pattern can be observed from any of the images after this point, the models also show very identical 'guessing' behavior (10%-15% correctness in classification).

### 3.6 Mathematical Modeling of the trend in accuracies shown by CNNs

In this section, we attempt to fit a mathematical equation to the trend followed by the considered FNN and CNN models.

#### 3.6.1 For an attack step size considered upto 16

| Attack step-size<=16 | Linear | Exponential | Logarithmic | Polynomial | Power Series |
|---|---|---|---|---|---|
| CNN(3 convolutions) | 0.112 | 0.066 | 0.509 | 0.264 | 0.513 |
| CNN(4 convolutions) | 0.189 | 0.209 | 0.613 | 0.325 | 0.744 |
| CNN(6 convolutions) | 0.123 | 0.07 | 0.541 | 0.267 | 0.49 |
| CNN(8 convolutions) | 0.409 | 0.482 | 0.834 | 0.531 | 0.879 |

*Table 8. Squared correlation values of the model fits for the decline of classification accuracies of networks for attack step sizes upto 16*

The squared correlation from the above table doesn't show many positively correlated model accuracies and nearest fits generated. The only two noticeable fits are:

CNN (8 - convolutions) which can be fit into the logarithmic equation,

$$f(x) = 0.524 - 0.149 \ln(x)$$

CNN (8 - convolutions) which can also be fit into the power series equation,

$$g(x) = 0.463 x^{-0.313}$$

Except for the above two equations, the others are not likely reliable owing to their small value of correlation. Thus, we attempt to further the study by modeling the data only upto an attack step size of 2 so that the classification accuracies of higher noise intensities that result only in the models 'guessing' the output are omitted.

**3.6.2 For an attack step size considered upto 2**

As mentioned in Section 3.4, we cannot expect a deviation trend from FNNs. We can approximately fit them into the following lines:

FNN (1 - hidden layer) is following a linear equation,

$$p(x) = 88.08$$

FNN (2 - hidden layer) is following a linear equation,

$$q(x) = 90.18$$

We attempted to fit the decline of classification accuracies of the networks against attack step sizes using linear, exponential, polynomial, logarithmic and power series equations.

| Attack step-size<=2 | Linear | Exponential | Logarithmic | Polynomial | Power Series |
|---|---|---|---|---|---|
| CNN(3 convolutions) | $-0.294x + 0.592$ | $0.757 e^{-1.02x}$ | $0.197 - 0.258 \ln x$ | $0.896 - 1.34x + 0.512x^2$ | $0.187 x^{-0.623}$ |
| CNN(4 convolutions) | $-0.283x + 0.577$ | $0.725 e^{-0.999x}$ | $0.2 - 0.245 \ln x$ | $0.854 - 1.23x + 0.466x^2$ | $0.19 x^{-0.612}$ |

| | Linear | Exponential | Logarithmic | Polynomial | Power Series |
|---|---|---|---|---|---|
| CNN(6 convolutions) | $-0.362x + 0.66$ | $0.858e^{-1.29x}$ | $0.189 - 0.294 \ln x$ | $0.973 - 1.44x + 0.528x^2$ | $0.166x^{-0.779}$ |
| CNN(8 convolutions) | $-0.332x + 0.853$ | $0.932e^{-0.662x}$ | $0.439 - 0.239 \ln x$ | $1.03 - 0.936x + 0.296x^2$ | $0.418x^{-0.414}$ |

**Table 9. Equations of the model fits for the decline of classification accuracies of networks for attack step sizes upto 2**

We tabulated the squared correlation ($r^2$) to understand the closeness of fits generated with the actual trend curves of the models.

| Attack step-size <=2 | Linear | Exponential | Logarithmic | Polynomial | Power Series |
|---|---|---|---|---|---|
| CNN(3 convolutions) | 0.427 | 0.607 | 0.774 | 0.794 | 0.766 |
| CNN(4 convolutions) | 0.455 | 0.638 | 0.802 | 0.805 | 0.813 |
| CNN(6 convolutions) | 0.566 | 0.803 | 0.881 | 0.907 | 0.908 |
| CNN(8 convolutions) | 0.773 | 0.879 | 0.941 | 0.947 | 0.919 |

**Table 10. Squared correlation values of the model fits for the decline of classification accuracies of networks for attack step sizes upto 2**

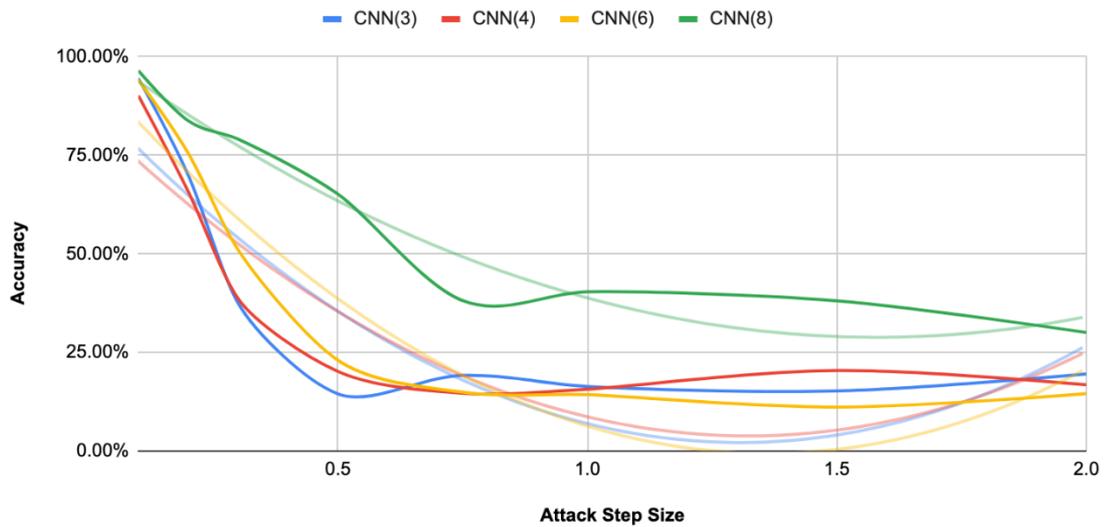

*Figure 1. Plot of decline in classification accuracies against the attack step sizes for the networks along with suitable polynomial fits*

The r² values for logarithmic, polynomial and power series fit are comparable and acceptable due to the presence of positive correlation between the model-fit generated and available data curve. Since the slope of logarithmic and power series models are asymptotic, the polynomial model is used for further study. It is important to note that the polynomial fits proposed are only applicable upto the minimum of the polynomial equations after which the value of the function is increasing. But in our case, the accuracy of the model does not increase in that pattern after it has come down. We can see that all the polynomial fits attain their minimum between at attack step size of 1 and 1.5. This goes in accordance to our observation in section 3.5 where any amount of noise beyond an attack step size of 1 did not make a big difference to the accuracy of the models and that point served as a breaking point of the accuracy of the models.

The below table shows the correlation values of the polynomial fits for the decline of accuracies of models with attack step size being less than or equal to 2. This gives us a clearer picture of the extent of correlation between the experimental results and computationally computed polynomial fits.

| Attack step-size <=2 | Correlation value |
|---|---|
| CNN(3 convolutions) | 0.891 |
| CNN(4 convolutions) | 0.897 |
| CNN(6 convolutions) | 0.952 |
| CNN(8 convolutions) | 0.973 |

*Table 11. r (correlation) for polynomial fits of the networks for attack step sizes upto 2*

The below equations represent the polynomial fits for CNNs with 3, 4, 6 and 8 convolutions respectively.

$$a(x) = 0.51x^2 - 1.34x + 0.896 \tag{1}$$

$$b(x) = 0.466x^2 - 1.23x + 0.854 \tag{2}$$

$$c(x) = 0.528x^2 - 1.44x + 0.973 \tag{3}$$

$$d(x) = 0.296x^2 - 0.936x + 1.03 \tag{4}$$

The obtained polynomial equations are differentiated to find the rate at which the classification accuracies of the models are declining.

$$a'(x) = 1.02x - 1.34 \tag{5}$$

$$b'(x) = 0.932x - 1.23 \tag{6}$$

$$c'(x) = 1.056x - 1.44 \tag{7}$$

$$d'(x) = 0.592x - 0.936 \tag{8}$$

The above functions are the slopes of the polynomial fits with x denoting the attack step size. They show the rate at which the accuracy declines. Differentiating it one more time would give us a value which denotes the slope of the rate of declination or the deceleration of accuracy when the attack step size increases.

In the below graph, we have plotted the lines that represent the deceleration of accuracies of models under adversarial noise and we see that the accuracy of CNN with 8 convolutions decelerates at a slower pace.

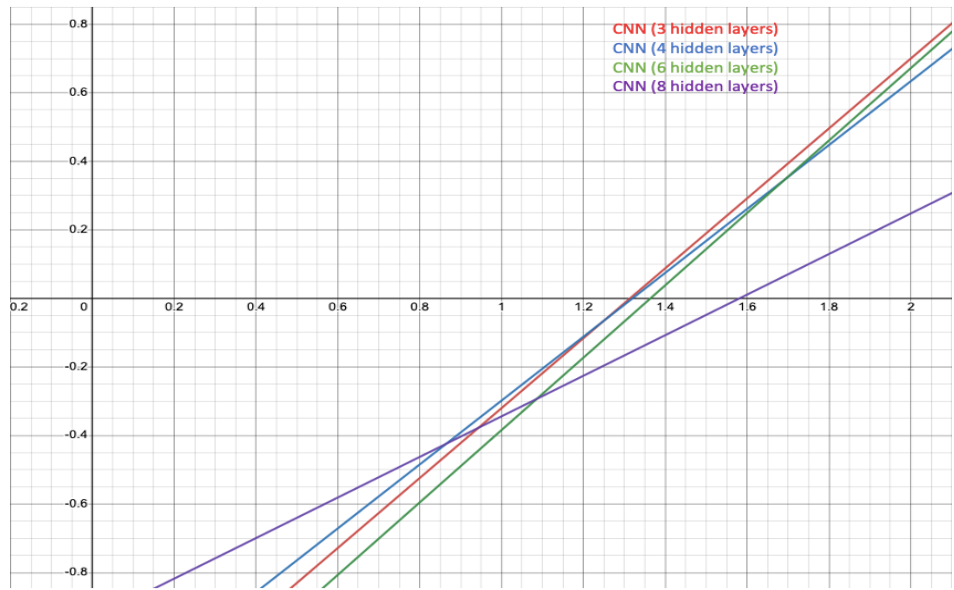

*Figure 2. Slopes of the polynomial fits that represent the decline of accuracies of the models for attack step sizes lesser than or equal to 2*

| Model | Rate of declination | Slope of rate of declination (deceleration of accuracy) |
|---|---|---|
| CNN (3 convolutions) | $1.02x - 1.34$ | 1.02 |
| CNN (4 convolutions) | $0.932x - 1.23$ | 0.932 |
| CNN (6 convolutions) | $1.056x - 1.44$ | 1.056 |
| CNN (8 convolutions) | $0.592x - 0.93$ | 0.592 |

***Table 12. Table denoting the rate of declination and the deceleration of accuracy for the CNNs developed under adversarial noise.***

We observe that the slope of the rate of declination of CNN with 8 convolutions is approximately half of that of the slope of rate of declination of CNNs with 3,4,6 convolutions.

CNN, with 8 convolutions has better accuracies even at higher attack step sizes when compared to other CNNs. This is corroborated by the value of the slope of rate of declination which is twice as much in the other CNNs as it is in the CNN with 8 convolutions.

We can observe that the line denoting the rate of declination of accuracy of CNN with 8 convolutions intersects with the lines denoting the rate of declination of accuracy of CNN with 3 and 4 convolutions between attack step size 0.8 and 1.0 and intersects the line denoting the rate of declination of accuracy of CNN with 6 convolutions between attack step size 1.0 and 1.2. At these points, the rate of declination of the corresponding models for adversarial data can be interpreted to be equal.

As discussed in Section 3.5 about attack step size of 1.0 acting as a breaking point, here it is also noticeable that it unifies the rate of declination of accuracy of all CNN models around itself.

## 3.7 Better performance of FNNs than CNNs on grayscale adversarial images

Recalling the architecture of FNNs, the relative positions between pixels is discarded while interpreting the relationships between the input and output vectors. Every pixel from input is individually analyzed and its significance is mapped using the adjusted weight matrix.

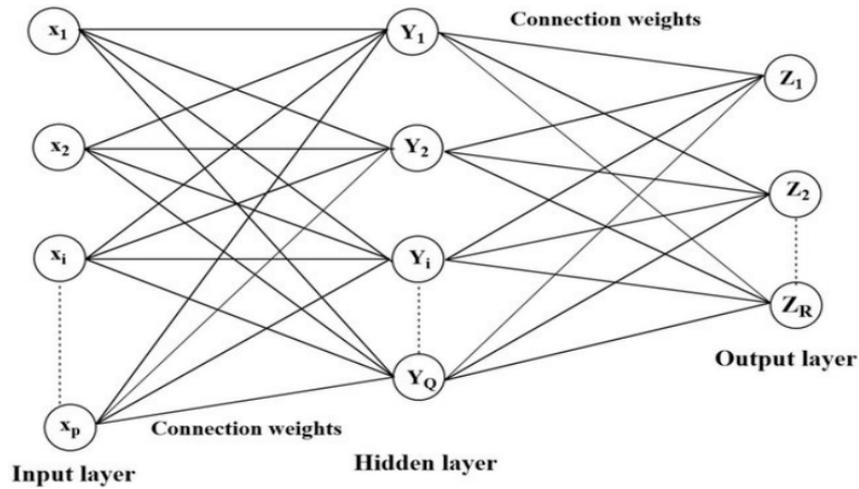

*Figure 3. A simple architecture of a Feedforward Neural Network [16]*

Preserving the local spatial coherence is a prominent principle of CNNs. Various filters extract features and this collection is condensed (max-pooled) a few times to extract higher-level features each time. We hypothesize that distorted features are presented to kernels while applying filters and when the convolutions are max-pooled, the noisy information gets accumulated and is carried through all the hierarchical layers. The distorted information of the image available at the highest feature extraction level is due to cumulative addition of inappropriate features right from lower-level.

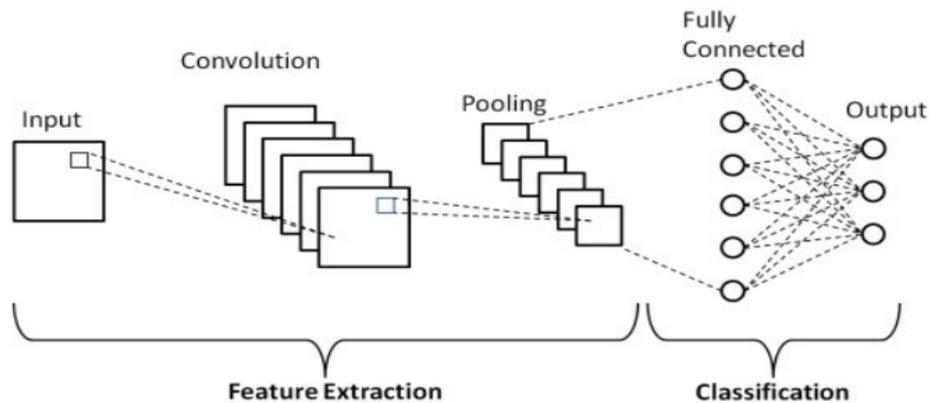

*Figure 4. A simple architecture of a Convolutional Neural Network [17]*

**Conclusion**

In this study, we have compared and contrasted the performance of CNNs and FNNs under adversarial noise for grayscale images and concluded that FNNs are much more robust to noise perturbation than CNNs. The reason for the same is also hypothesized by carefully considering the architecture of the models. The correlation between the trend of decline of accuracy among the models is also considered to comment on the similarity in behavior of models when the images are subjected to noise perturbation. The trends of decline in accuracy were captured using different mathematical models and the most suitable among them were employed. Using the models that approximate the real trend of accuracies, we attempted to understand the rate at which the accuracy drops for each of the models and qualitatively commented on which model exhibits relatively better robustness towards adversarial noise. This helped us to gain a better understanding as to for which concentration of noise the models behave similarly and the breaking point of the CNNs under adversarial noise.

**Future Scope**

An interesting problem statement to explore would be to understand how FNNs are able to classify the images of the MNIST dataset with an appreciable level of accuracy when the attack step size increases beyond 1 and no substantial pattern is decipherable to the human eye.